# Tactical Decision Making for Emergency Vehicles Based on A Combinational Learning Method

Haoyi Niu[1]; Jianming Hu[2]†; Zheyu Cui[3]; and Yi Zhang[4]

[1]Dept. of Automation, Tsinghua Univ., Beijing 100084, China. Email: niuhy18@mails.tsinghua.edu.cn

[2] (†Corresponding Author) Dept. of Automation, Tsinghua Univ., Beijing 100084, China; Tsinghua University (Department of Automation) - Shanghai Cleartv Co., Ltd. Joint Research Center for Video-based Scenario Fusion Technology. Email: hujm@mail.tsinghua.edu.cn

[3]Dept. of Automation, Tsinghua Univ., Beijing 100084, China. Email: czyfaiz@foxmail.com

[4] Dept. of Automation, Tsinghua Univ., Beijing 100084, China; Tsinghua University (Department of Automation) - Shanghai Cleartv Co., Ltd. Joint Research Center for Video-based Scenario Fusion Technology. Email: zhyi@mail.tsinghua.edu.cn

**ABSTRACT**

Increasing the response time of emergency vehicles (EVs) could lead to an immeasurable loss of property and life. On this account, tactical decision making for EVs' microscopic control remains an indispensable issue to be improved. In this paper, a rule-based avoiding strategy (AS) is devised, that common vehicles (CVs) in the prioritized zone ahead of EV should accelerate or change their lane to avoid it. Besides, a novel DQN method with speed-adaptive compact state space (SC-DQN) is put forward to fit in EVs' high-speed features and generalize in various road topologies. Afterward, the execution of AS feedback to the input of SC-DQN so that they joint organically as a combinational method. The following approach reveals that deep reinforcement learning (DRL) could complement rule-based AS in generalization, and on the contrary, the rule-based AS could complement the stability of DRL, and their combination could lead to less response time, lower collision rate, and smoother trajectory.

## I. INTRODUCTION

Emergency Technology Company RapidSOS mentioned a set of data: "In medical emergencies such as cardiac arrest, every one-minute delay in response time causes mortality rate to increase by 1% and imposes additional $1542 in hospital costs, leading to 7 billion dollars increase in healthcare expenditure per year only in the USA.

Similarly, one minute of reduced response time leads to a decrease in healthcare costs by 326,000 baht ($10,190) in Thailand."(RAPIDSOS, 2015) from which this paper considers cutting down the response time of EVs an unavoidable issue.

Most researches focus on route optimization and traffic signal preemption, which address the problem from a macroscopic perspective. However, real-time traffic data are not taken full advantage of. Besides, hardly has an impact on normal traffic been considered. Moreover, these deterministic strategies are harder to generalize to various traffic scenarios than policies acquired by DRL. Even so, It is not to say that autonomous vehicles could abandon rule-based policies since they outperform other methods in outstanding stability. Consequently, this model combines rule-based strategy and DRL method. For CVs, this model devises a priority zone where they have to meet safety requirements and give way to EVs simultaneously. For EVs, the DQN method enables them to change lanes actively on their own when CVs could not execute the AS on account of the minimum longitudinal gap requirement.

Our departure point is to shorten the response time of emergency vehicles in a highway environment, where limited research has been done. This paper tends to propose a framework with a time-saving, real-time, and data-efficient tactical decision-making method for EVs, and a rule-based AS for CVs.

The remainder of this paper is organized as follows: Section II gives a brief overview of the research on different aspects to cut down the response time of EVs. A detailed description of how we model the framework and how it performs follows in Section III and Section IV respectively. Eventually, we draw a conclusion and put forward considerable future work in Section V.

## II. RELATED WORK

Prior researches on shortening EV response time on highways mostly concentrate on traffic signal control, route optimization, and deterministic decision-making strategies.

### A. Traffic Signal Control

Tactical control at signalized intersections can assist EVs to pass with a priority, which is also known as Route-preemption. The most commonly used method is controlling traffic signals at the intersection where EV is upcoming, to halt lower-prioritized traffic(Mu, Song and Liu, 2018) or disperse the occluded traffic in advance(Noori, Fu and Shiravi, 2016), and eventually reduce the delay of EVs. In (Kang *et al.*, 2014), a method that takes the coordination of intersections into consideration is mentioned, with a less negative impact on the normal traffic streams.

### B. Route Optimization

It's very significant to dispatch EVs in optimal routes to keep away from traffic congestion. Dynamic route guidance systems(Gong, Yang and Lin, 2009) and vehicle evacuation algorithm(Gong, Yang and Lin, 2009) based on Dijkstra have been devised, and other path optimization algorithms like A* and Ant Colony Algorithm are also summarized in an overview(Humagain *et al.*, 2020). In (Zhao, Guo and Duan, 2017), A two-stage shortest path algorithm composed of K-paths algorithm and shuffled frog leaping algorithm is proposed. (Brady and Park, 2016) demonstrates emergency vehicle route guidance by utilizing additional road features. All above is about optimizing the response time of EV from a macroscopic perspective, so real-time data is usually not leveraged to the utmost like what microscopic control tactic does.

### C. Rule-based Decision-making Policy

Limited researches have shown concern to microscopic control of EVs. A series of EV traversal models are mentioned in (Ismath *et al.*, 2019), incorporating Fixed Lane Strategy, Best Lane Strategy, Lane Change Algorithm, and Risk Assessment Algorithm. From Fixed Lane Strategy, they construct a priority zone and CVs have to comply with an avoiding policy if they block the way of EV within the priority distance. That is not to say deterministic policies cannot perform well in certain scenarios, but they do not have many advantages in exploration for lane changes and adaptation for changeable road topologies against Deep Reinforcement Learning methods.

To summarize, most technologies and strategies for EVs are macroscopic and traffic-signal-based while approaches for microscopic control in the highway domain are few and far between.

However, response time spent on the highway should not be overlooked, especially where the distance between adjacent junctions also contributes crucially to the total traversal time. Although rule-based AS could guarantee the stability of EVs' performance, it could hardly generalize to find optimal policies in different traffic scenarios. Fortunately, DRL algorithms have already become a powerful learning framework capable of figuring out this problem. This survey(Kiran *et al.*, 2020) comprises a systematical taxonomy of autonomous driving, but few researches implement DRL in how to control EVs.

Accordingly, this method innovatively fits modified DQN in a semantic environment for EVs while CVs follow AS, trying to acquire a method with both stability and optimality in this combinational circumstance.

### III. TASK MODELING

The framework is implemented in SUMO(Bieker-Walz *et al.*, 2011) traffic emulator, with DRL method for EVs and rule-based AS for CVs. First and foremost, we set up the speed-adaptive compact state space and discrete action space of the RL agent. Afterward, EV-specific reward functions are devised. Finally, an introduction about how this model initializes the input of the environment and how we structure the Speed-adaptive Compact Deep Q-Network (SC-DQN: an ad-hoc DQN method for autonomous EV) is going to be discussed.

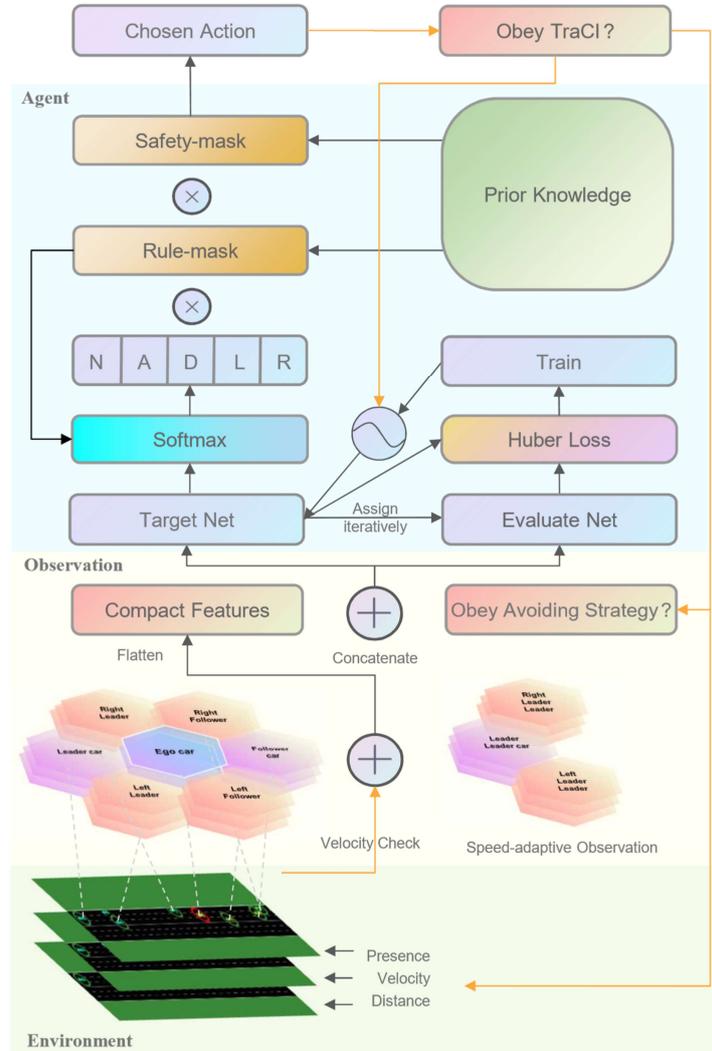

**Figure 1. Algorithm Architecture for SC-DQN in Section Ⅲ.**

*A. Speed-adaptive Compact State Space*

Other previous work about compact state space (Peter Wolf, Karl Kurzer, Tobias Wingert, 2018) has referred to a semantic method to structure a relational grid about neighboring vehicles. To make the learning process more data-efficient, this method refines the state space in a more compact way (Figure 2), which originally contains 6 neighbors of EV (ego car). Moreover, this kind of compact structure could fit in variant road topologies, including but not limited to curve, ramp, and merge. The state space extracts three features of each neighbor: *Presence* is whether its neighbor exists. *Velocity* and *Distance* of this neighbor is zero when *Presence* is assigned 'True'. It is noteworthy that Figure 2 only demonstrates the directional relation of EV and its neighbors, but does not depict that of EVs' neighbors (i.e. Left Leader may not be the Left Follower of Leader Car (Figure 2)).

Intuitively, auxiliary speed-adaptive features should be additive to the EV compact observation for a more forward-looking view if the velocity of EV is more than half of its maximum velocity. These incorporate leader of leader car, left leader car, and right leader car, called speed-adaptive observation in Figure 1.

Speed-adaptive observation is concatenated with normal compact observation, together with the execution effect of AS (mentioned in (Ismath *et al.*, 2019), which denotes that leader car ought to avoid the EV within the priority distance). Whether the leader car executes the AS successfully is a key feedback feature that plays an important role in this model, and also a crucial innovation point for combining the DRL algorithm with the rule-based strategy organically, the simulation result of which will be presented and analyzed in the Section IV.

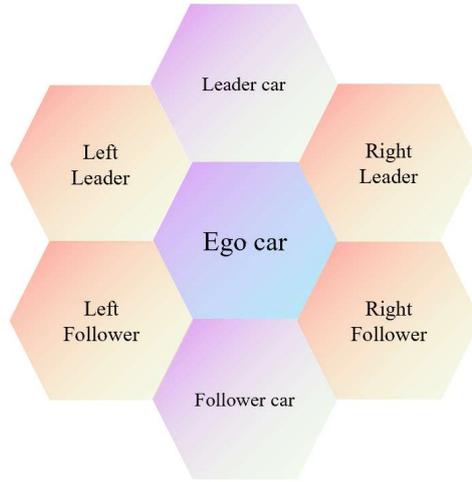

**Figure 2. Compact Neighbors**

## B. Discrete Action Space

In the lateral direction, there are three choices for the ego car: lane-keeping, lane-changing to left, and lane-changing to right. In addition, from the longitudinal dimension, the ego car could accelerate or decelerate at a certain acceleration in the current simulation step. Mapping from the two dimensions, action space consists of 5 elements:

- *N*  no-operation;
- *A*  longitudinally accelerate at $3m/s^2$ in this simulation step;
- *D*  longitudinally decelerate at $3m/s^2$ in this simulation step;
- *L*  makes a left lane-change;
- *R*  makes a right lane-change;

## C. Reward Designing

To evaluate the reasonability of chosen actions and guide the training agent into a convergence, reward designing for autonomous driving basically focuses on four aspects($r_{col}$, $r_v$, $r_{lc}$, $r_{cor}$) as follows.

- **Safety**

    Unlike the ordinary way that gives a negative reward when a collision occurs, this model implements a continuous reward function proportional to survival distance, serving as a real-time stimulus:

$$r_{col} = \frac{d_{survive} - d_{total}/2}{d_{total}} \quad (1)$$

$d_{total}$ is the total distance of this section of highway; $d_{survive}$ is the current distance of safely driving.

- **Efficiency**

    Efficiency reward is also proportional to the current velocity, following the continuous reward designing concept as what we do in 'Safety'.

$$r_v = \frac{v_{current} - v_{max}/2}{v_{current} - v_{min}} \quad (2)$$

$v_{current}$ is the current velocity of EV; $v_{max}$ and $v_{min}$ are the allowed maximum and minimum velocity of EV respectively.

- **Smoothness**

    RL-based autonomous vehicles have always been denounced for their undesirable lane-changing frequency. Hence, it is intuitive to consider lane-changing at EVs' maximum velocity a meaningless behavior because lane-changing is intuitively an action for pursuing higher velocity, but it could not obtain higher velocity by lane-changing at that time. Thus, $r_{lc}$(=-1) will be given if changing lane at its maximum velocity, to smooth the trajectory and alleviate the impact of frequent lane-changes to normal traffic, and surely assist the learning process more convergent.

- **Cooperation***

    In Section IV, the preliminary results show that the combination of SC-DQN and AS underperforms against only AS. That's because SC-DQN produces excessive lane-changes that bring about disorder and congestion to normal traffic, which later has an inverse impact on EVs. Thus, EVs ought to take account of the traversal efficiency of CVs ahead of it, like what equation 3 tells, to weight leading CVs' speed in an exponentially decaying way, by which EVs tend to keep the traffic lighter. (This section is exclusive for "SC-DQN + AS (with Cooperative Reward)" in Table 3, and not included in SC-DQN)

$$r_{cor} = \frac{1}{i} \sum_{i=0}^{n} \frac{v_i}{v_{max}^j} e^{-i} \qquad (3)$$

Among equation 3, n is the number of CVs in the priority zone of EV; $v_i$ and $v_{max}^j$ are the current velocity and allowed maximum velocity of the ith CV.

The total reward for each step is accumulated as equation 4:

$$R_t = r_{col} + r_v + r_{lc} + r_{cor} \qquad (4)$$

### *D. Mask of Prior Knowledge*

A multitude of researchers adopts prior knowledge in their RL environments, addressing the problem that RL methods show a natural instability against some proven policies. Additionally, as a preprocessing, prior knowledge initializes the agent at the very beginning, so fewer episodes are spent learning road constraints and traffic rules.

Consequently, this model sets up two masks to constrain the output action. One is Rule Mask, to check whether the action that maximizes Q value complies with the traffic rules and road topology constraints. For instance, if EV takes a right lane change on the rightmost lane or accelerates to a velocity that surpasses the maximum velocity, the invalid action will be rejected and the suboptimal action according to Q values will be selected. This is similar to the Q-masking technique in (Mukadam *et al.*, 2017). Another is Safety Mask. RL-based methods are well-known for the lane-changing initiative, whereas safety issue is always the drawback that hardly converges. In SUMO, this model initializes the EVs with the parameter laneChangeMode=512 (disables all autonomous changing but still handle safety checks in the simulation, either one of the modes 256 (collision avoidance) or 512 (collision avoidance and safety-gap enforcement) may be used.(Erdmann, 2015)). Thus, the lane change mode blocks the safety-challenging actions and chooses one from its experience. If TraCI, the configurable Python interface for SUMO, could not execute what SC-DQN commands, this step will not be learned by the network.

### *E. Network Architecture*

Deep Q-Network (Mnih *et al.*, 2015) is one of the most practical RL-based network architectures that has been implemented in autonomous driving. In this model, transformed DQN utilizes speed-adaptive state space and the execution of AS as its current observation($O_t$). The target network contains an input layer of 32 neurons, two hidden layers with 20 and 10 neurons respectively, and the output layer with 5 neurons for each action.

With the probability of exploration rate $\epsilon$(initialized as 0.9 and anneals $4 \times 10^{-6}$ each step until reaching 0.1), Action($A_t$) is selected by Q-values, the output of the target network, otherwise, action will be chosen randomly as an exploration. Every 5000 iterations, parameters of the target network are assigned to those of the evaluate network, a network with the same structure, which is a crucial step leading to the final convergence.

Successor state $O_{t+1}$ and reward $R_t$ follow the interaction between $A_t$ and the environment. $(O_t, A_t, R_t, O_{t+1})$ is a transition and it will be stored in the memory pool with the capacity of 2000 transitions, if $A_t$ is determined by RL-agent instead of prior knowledge.

Hereafter, SC-DQN (Speed-adaptive Compact Deep Q-Network) will be trained with experience replay technique(Schaul *et al.*, 2015) and update the mini-batch with 32 transitions every learning step. Furthermore, the Q value of Evaluate Network and Target Network are calculated as equation 5 and 6. Besides, Adam Optimizer is applied with the learning rate decaying exponentially. Last but not the least, to prevent the phenomenon of gradient explosion, Deep Q-Network learns its parameters to minimize Huber Loss ($L_H(\theta)$ in equation 7) instead of Mean-Square Loss:

$$Q = Q(O_t, A_t; \theta) \tag{5}$$

$$Q' = R_t + \gamma \max_{A_{t+1}} Q(O_{t+1}, A_{t+1}; \theta^-) \tag{6}$$

$$L_H(\theta) = \begin{cases} \frac{1}{2}(Q'-Q)^2 & \text{for } |Q'-Q|<1 \\ |Q'-Q|-\frac{1}{2} & \text{otherwise} \end{cases} \tag{7}$$

## IV. EVALUATION

On a 2000-meter highway for simulation test, we implement this model with random normal traffic flow and parameter-setup EVs. Table 1 and 2 are non-default parameters setup for EVs and CVs:

**TABLE 1. HOW EVS PARAMETERS ARE SET UP**

| *Length/m* | *Width/m* | *minGap/m* |
|---|---|---|
| 5.0 | 2.0 | 1.0 |
| *maxSpeed/m/s* | *accel/m/s²* | *decel/ m/s²* |
| 40.0 | 4.0 | 4.0 |
| *lcKeepRight* | *lcSublane* | *lcStrategic* |
| 0.0 | 0.0 | 0.0 |

**TABLE 2. HOW CVS PARAMETERS ARE SET UP**

| *Length/m* | *Width/m* | *minGap/m* |
|---|---|---|
| 5.0 | 1.8 | 2.0 |
| *maxSpeed/m/s* | *accel/m/s²* | *decel/ m/s²* |
| 20.0 | 2.0 | 2.0 |
| *lcKeepRight* | *sigma* | *lcPushy* |
| 0.0 | 1.0 | 0.5 |

First and foremost, the loss function of SC-DQN performs in a convergence (in Figure 3(Left) after nearly 200 thousand training episodes.

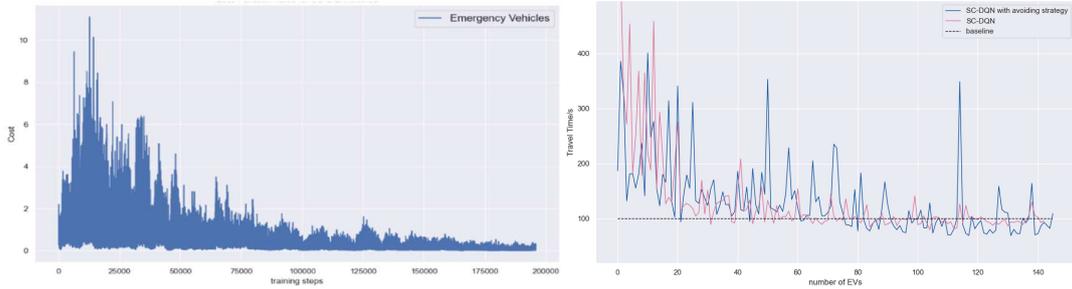

**Figure 3. The convergence of Loss Function (Left) and Response Time (Right).**

Whereafter, from the preliminary work came a suite of discoveries leading to the drawbacks of AS in (Ismath *et al.*, 2019) and SC-DQN method respectively. It reveals that AS for CVs at a low velocity and low traffic area benefits EVs a lot, however, it is proved deadlock-prone when at a high velocity and high traffic area because CVs are more likely to address their safety than EVs' efficiency. Intuitively, SC-DQN is no more advantages if leading CVs are bound to avoid for EVs while it could mitigate the conflict between CVs' concern for safe distance and EVs' desire for higher speed, so it outperforms in high-speed traffic flow. As shown in Figure 3(Right), *SC-DQN* and *SC-DQN with AS feedback* both converge to better performance than the baseline.

Based on the above discussion, this paper is going to compare *baseline*, *SC-DQN*, *Avoiding Strategy*, *SC-DQN with Avoiding Strategy,* and *the combinational method with modified reward function* in the simulation: **1)**The baseline is a method with prior knowledge based on a default lane-keeping strategy in SUMO, which could barely meet our standards; **2)**SC-DQN is speed-adaptive compact DQN, an ad-hoc DQN method that could fit well in an environment with EVs and different road topologies; **3)**Avoiding Strategy serves as a benchmark based on the method mentioned in (Ismath *et al.*, 2019); **4)**An organic combination of SC-DQN for EVs and Avoiding Strategy for CVs in which CVs' execution of Avoiding Strategy serves as a feedback in the state space of RL agent; **5)** Append a cooperation item to the reward function, based on the method mentioned in 4).

From Table 3, this paper evaluates these algorithms in three main aspects: efficiency, safety, and smoothness. The combination method of SC-DQN and Avoiding Strategy, compared to SC-DQN, facilitates the agent to pursue higher velocity and lower collision rate, but the high lane-changing frequency is incomparable to that of human drivers. Besides, from the perspective of efficiency, the methods above could not outperform the Avoiding Strategy. With an auxiliary cooperative reward mentioned in Section Ⅲ C, the combinational method could obtain less traversal time with a lower lane-change frequency and also behave well in collision rate. In Figure 4, we record the time-distance curve of EVs and CVs, which apparently reveals that the cooperative reward enables the EVs less frequent to encroach on the right of way of others, and the traffic more efficient and orderly.

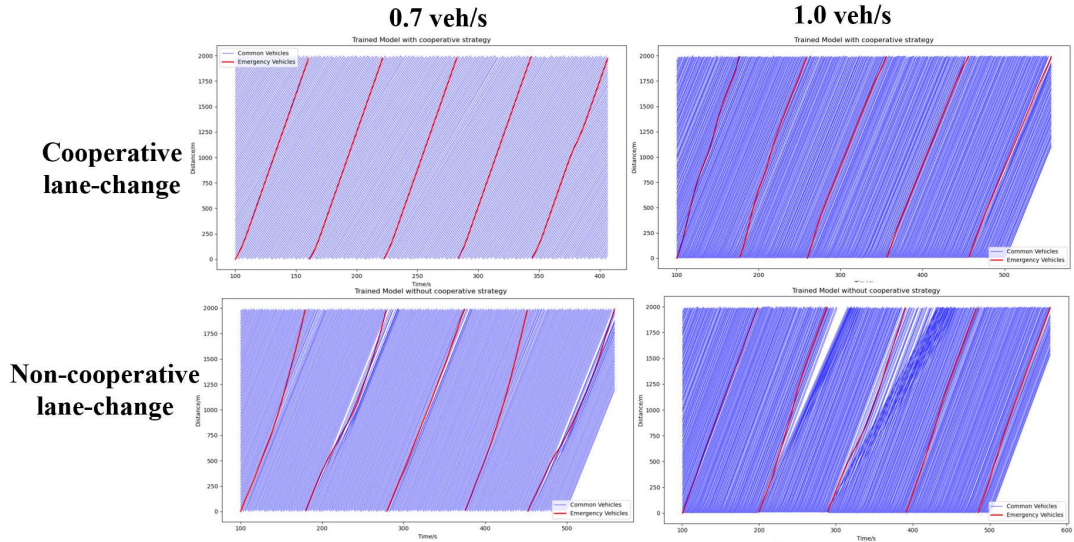

**Figure 4. Append cooperative reward before and after.**

**TABLE 3. SIMULATION RESULTS OF OUR METHOD AGAINST OTHERS**

| Method | Traffic flow (Avg input vehicles/s) | Evaluation | | |
| --- | --- | --- | --- | --- |
| | | Avg Travel Time/s | Collision Rate/% | Avg Lane Changes /(2km)$^{-1}$ |
| Baseline | 0.5 | 100.1 | 80.0 | 0 |
| | 1 | 105.2 | 83.9 | 0 |
| SC-DQN | 0.5 | 93.3 | 0.0 | 130 |
| | 1 | 101.1 | 0.0 | 39.3 |
| Avoiding Strategy | 0.5 | 69.0 | 0.7 | 0 |
| | 1 | 94.9 | 46.7 | 0 |
| SC-DQN+Avoiding Strategy | 0.5 | 70.7 | 0.0 | 55.6 |
| | 1 | 91.9 | 0.0 | 7.75 |
| SC-DQN+Avoiding Strategy (with Cooperative Reward*) | 0.5 | 62.5 | 0.0 | 0.02 |
| | 1 | 91.1 | 0.0 | 6.67 |

## V. CONCLUSION

In this work, we have devised a framework about how autonomous EVs in self-driving traffic flow make tactical decisions based on DQN with a speed-adaptive and data-efficient observation. For the first time, the idea of combining rule-based AS with DRL algorithms organically is proposed, which allows the execution of AS to feed back for the DRL observation. It bridges the gap between deterministic policies and RL-based algorithms in a complementary way. But it is not to say that this model has solved the instability of DQN, because it utilizes some prior knowledge to mask off excessive lane-changes to smooth the trajectory, and we are still exploring how to ameliorate the trajectory smoothness and narrow the Sim-to-Real gap by methods like RL with intrinsic reward (Zheng, Oh and Singh, 2018b)(Zheng, Oh and Singh, 2018a), Model-based RL(Curi, Berkenkamp and Krause, 2020) and Attention-based Hierarchical RL(Yilun Chen, Chiyu Dong, Praveen Palanisamy, Priyantha Mudalige, 2019). Additionally, absolute congestions are neglected in our modeling hypothesis, so finding out a congestion-free Sublane-changing Model based on microscopic simulation framework(van Lint and Calvert, 2018)(Calvert and van Arem, 2020)(Bieker-Walz, Behrisch and Junghans, 2018) will also be a considerable choice in future work.

## ACKNOWLEDGMENT

This work is supported by National Key R&D Program in China (2019YFF0303102), National Natural Science Foundation of China under Grant No. 61673232, and Tsinghua University Initiative Scientific Research Program (2018Z05JDX005, 20183080016).

## REFERENCES

Bieker-Walz, L. *et al.* (2011) *SUMO-Simulation of Urban MObility: An Overview SUMO-Simulation of Urban MObility An Overview*. Available at: https://www.researchgate.net/publication/225022282.

Bieker-Walz, L., Behrisch, M. and Junghans, M. (2018) 'Analysis of the traffic behavior of emergency vehicles in a microscopic traffic simulation', in. EasyChair, pp. 1--13. doi: 10.29007/bv4j.

Brady, D. and Park, B. B. (2016) 'Improving emergency vehicle routing with additional road features', *VEHITS 2016 - 2nd International Conference on Vehicle Technology and Intelligent Transport Systems, Proceedings*, (Vehits), pp. 187–194. doi: 10.5220/0005844701870194.

Calvert, S. C. and van Arem, B. (2020) 'A generic multi-level framework for microscopic traffic simulation with automated vehicles in mixed traffic', *Transportation Research Part C: Emerging Technologies*. Elsevier Ltd, 110, pp. 291–311. doi: 10.1016/j.trc.2019.11.019.

Curi, S., Berkenkamp, F. and Krause, A. (2020) 'Efficient Model-Based Reinforcement Learning through Optimistic Policy Search and Planning'. Available at: http://arxiv.org/abs/2006.08684.

Erdmann, J. (2015) 'SUMO's Lane-changing model', *Lecture Notes in Control and Information Sciences*, 13, pp. 105–123. doi: 10.1007/978-3-319-15024-6_7.

Gong, B., Yang, Z. and Lin, C. (2009) 'Dispatching optimization and routing guidance for emergency vehicles in disaster', in *Proceedings of the 2009 IEEE International Conference on Automation and Logistics, ICAL 2009*, pp. 1121–1126. doi: 10.1109/ICAL.2009.5262588.

Humagain, S. *et al.* (2020) 'A systematic review of route optimisation and pre-emption methods for emergency vehicles', *Transport Reviews*. Routledge, 40(1), pp. 35–53. doi: 10.1080/01441647.2019.1649319.

Ismath, I. *et al.* (2019) 'Emergency Vehicle Traversal using DSRC/WAVE based Vehicular Communication IEEE Intelligent Vehicles Symposium (IV).', in *IEEE Intelligent Vehicles*

*Symposium (IV)*.

Kang, W. *et al.* (2014) 'Traffic signal coordination for emergency vehicles', *2014 17th IEEE International Conference on Intelligent Transportation Systems, ITSC 2014*, (October), pp. 157–161. doi: 10.1109/ITSC.2014.6957683.

Kiran, B. R. *et al.* (2020) 'Deep Reinforcement Learning for Autonomous Driving: A Survey', (February). Available at: http://arxiv.org/abs/2002.00444.

van Lint, J. W. C. and Calvert, S. C. (2018) 'A generic multi-level framework for microscopic traffic simulation—Theory and an example case in modelling driver distraction', *Transportation Research Part B: Methodological*. Elsevier Ltd, 117, pp. 63–86. doi: 10.1016/j.trb.2018.08.009.

Mnih, V. *et al.* (2015) 'Human-level control through deep reinforcement learning', *Nature*, 518(7540), pp. 529–533. doi: 10.1038/nature14236.

Mu, H., Song, Y. and Liu, L. (2018) 'Route-Based Signal Preemption Control of Emergency Vehicle', *Journal of Control Science and Engineering*, 2018. doi: 10.1155/2018/1024382.

Mukadam, M. *et al.* (2017) 'Tactical Decision Making for Lane Changing with Deep Reinforcement Learning', in. Long Beach, CA, USA: 31st Conference on Neural Information Processing Systems (NIPS 2017).

Noori, H., Fu, L. and Shiravi, S. (2016) 'A Connected Vehicle Based Traffic Signal Control Strategy for Emergency Vehicle Preemption', *TRB,95th Annual Meeting of the Transportation Research Board*, (May).

Peter Wolf, Karl Kurzer, Tobias Wingert, F. K. and J. M. Z. (2018) 'Adaptive Behavior Generation for Autonomous Driving using Deep Reinforcement Learning with Compact Semantic States', in. Changshu, p. 993.

RAPIDSOS (2015) 'Outcomes Quantifying the Impact of Emergency Response Times', (0), pp. 3–4. Available at: https://cdn2.hubspot.net/hubfs/549701/Documents/RapidSOS_Outcomes_White_Paper_-_2015_4.pdf.

Schaul, T. *et al.* (2015) 'Prioritized Experience Replay', *arXiv: Learning*.

Yilun Chen, Chiyu Dong, Praveen Palanisamy, Priyantha Mudalige, K. M. and J. M. D. (2019) 'Attention-based Hierarchical Deep Reinforcement Learning for Lane Change Behaviors in Autonomous Driving', in. Macau, China, pp. 3697–3703.

Zhao, J., Guo, Y. and Duan, X. (2017) 'Dynamic Path Planning of Emergency Vehicles Based on Travel Time Prediction', *Journal of Advanced Transportation*, 2017. doi: 10.1155/2017/9184891.

Zheng, Z., Oh, J. and Singh, S. (2018a) 'On learning intrinsic rewards for policy gradient methods', *Advances in Neural Information Processing Systems*, 2018-Decem, pp. 4644–4654.

Zheng, Z., Oh, J. and Singh, S. (2018b) 'On Learning Intrinsic Rewards for Policy Gradient Methods'. Available at: http://arxiv.org/abs/1804.06459.